# LLMs for Low-Resource Dialect Translation Using Context-Aware Prompting: A Case Study on Sylheti

Tabia Tanzin Prama[1,2,3,5], Christopher M. Danforth[1,2,3,4], Peter Sheridan Dodds[1,2,3,5,6]

[1]Computational Story Lab, [2]Vermont Complex Systems Institute,
[3]Vermont Advanced Computing Center,
[4]Department of Mathematics and Statistics, [5]Department of Computer Science,
University of Vermont, Burlington, VT 05405, USA
[6]Santa Fe Institute, 1399 Hyde Park Rd, Santa Fe, NM 87501, USA

## Abstract

Large Language Models (LLMs) have demonstrated strong translation abilities through prompting, even without task-specific training. However, their effectiveness in dialectal and low-resource contexts remains underexplored. This study presents the first systematic investigation of LLM-based Machine Translation (MT) for Sylheti, a dialect of Bangla that is itself low-resource. We evaluate five advanced LLMs (GPT-4.1, GPT-4.1, LLaMA 4, Grok 3, and Deepseek V3.2) across both translation directions (Bangla ⇔ Sylheti), and find that these models struggle with dialect-specific vocabulary. To address this, we introduce Sylheti-CAP (Context-Aware Prompting), a three-step framework that embeds a linguistic rulebook, dictionary (2260 core vocabulary and idioms), and authenticity check directly into prompts. Extensive experiments show that Sylheti-CAP consistently improves translation quality across models and prompting strategies. Both automatic metrics and human evaluations confirm its effectiveness, while qualitative analysis reveals notable reductions in hallucinations, ambiguities, and awkward phrasing—establishing Sylheti-CAP as a scalable solution for dialectal and low-resource MT. Dataset link: https://github.com/Sylheti-CAP

## 1 Introduction

Large Language Models (LLMs) have recently demonstrated remarkable potential in natural language processing (NLP) tasks (Yang et al., 2024; Dubey et al., 2024; OpenAI et al., 2023), including neural machine translation (NMT). Prior studies (Robinson et al., 2023; Zhu et al., 2023) show that while LLMs achieve strong performance in translating high-resource languages, their effectiveness decreases significantly for low-resource languages (LRLs) (Joulin et al., 2016; team et al., 2022), where parallel data is limited and difficult to obtain.

Compared to traditional NMT models, LLMs offer several qualitative advantages. They allow controllability of style and language variety through prompting and in-context learning (Brown et al., 2020; García et al., 2023; Agrawal et al., 2022), exhibit inherent document-level translation capabilities (Wang et al., 2023; Karpinska and Iyyer, 2023), produce less literal translations (Raunak et al., 2023), and demonstrate improved handling of complex linguistic phenomena such as idioms and ambiguous expressions. Consequently, LLMs are increasingly surpassing conventional NMT models in versatility (Peng et al., 2023; Hendy et al., 2023; Zhu et al., 2023).

Recent research has leveraged in-context learning (ICL) (Brown et al., 2020; Dong et al., 2022) to enable LLMs to perform translation without parameter updates, and supervised fine-tuning with parallel corpora has also been explored (Li et al., 2023; Chen et al., 2021; Alves et al., 2023). However, training LLMs still requires vast multilingual resources, and the inherent imbalance in language coverage continues to hinder performance for many LRLs (Jiao et al., 2023; Hendy et al., 2023). While prior work has shown impressive results in high-resource pairs such as English–German translation (Vilar et al., 2022), the effectiveness of LLMs in dialect-specific scenarios remains underexplored.

This gap is particularly acute for languages like Bangla (Prama et al., 2025). More than two hundred million people speak Bangla (also known as Bengali) (Accredited Language Services, 2015), yet it remains relatively low-resource in the NLP landscape. Its dialects are even more underserved, with virtually no large-scale datasets. These dialects encode rich linguistic and cultural variation, but unlike the standardized language, they rarely benefit from curated resources such as newswire corpora. Sylheti is a major Bangla dialect with an estimated 11 million speakers worldwide (Simard et al., 2020), illustrates this problem especially clearly. Although a few studies have explored Bangla⇔Sylheti translation using traditional deep learning models (Prama and Anwar, 2025a; Faria et al., 2023), research remains limited. To our knowledge, this is the first systematic evaluation of LLM-based machine translation for Bangla ⇔ Sylheti. We frame our study around two research questions (RQs):

**RQ1: How do LLMs perform MT between Bangla and the Sylheti dialect?**

To answer this question, we evaluate multilingual LLMs (LLaMA-4 (AI, 2024), Gemini 2.5 Flash (DeepMind, 2025), GPT-4.1 (OpenAI, 2024), DeepSeek v3.2 (DeepSeek-AI, 2024), and Grok 3 (xAI, 2025)) from five different LLM families. Here the LLMs are first used in a zero-shot setting, meaning that we assume that (to the best of our knowledge) the models are not directly trained with Sylheti-specific data but are instead expected to apply their knowledge of Bangla to understand and translate Sylheti. On average, Sylheti → Bangla translation achieves 66.8 % higher BLEU-1 scores than Bangla→ Sylheti. Also Llama 4 and Grok achive superior perfromance among the models we tested.

**RQ2: How can we improve LLM translation performance?**

To address this question, we propose Sylheti-CAP, a context-aware prompting strategy designed to enhance LLM translation for low-resource dialects shows in Figure 1. While prior work has explored adding extra-sentential context to translation (Maruf et al., 2019; Castilho and Knowles, 2024), such models—trained solely for translation—have shown only modest gains over context-agnostic baselines (Chatterjee et al., 2020; Yin et al., 2021). Recent studies show that LLMs can effectively leverage contextual information for various NLP tasks, including document-level translation (Karpinska and Iyyer, 2023; Wang et al., 2023). Building on this, Sylheti-CAP integrates Sylheti-specific lexical, grammatical, and idiomatic knowledge (including untranslatable terms) directly into prompts, followed by a fluency and correctness refinement step. We evaluate Sylheti-CAP on Bangla ⇔ Sylheti translation using five LLMs. Results on BLEU, METEOR, and ChrF show consistent improvements over Zero-Shot, Few-Shot, and CoT prompting, with fewer mistranslations, omissions, and awkward phrases. Human preference and MQM (Lommel, 2013) evaluations further confirm that Sylheti-CAP yields more natural and faithful translations.

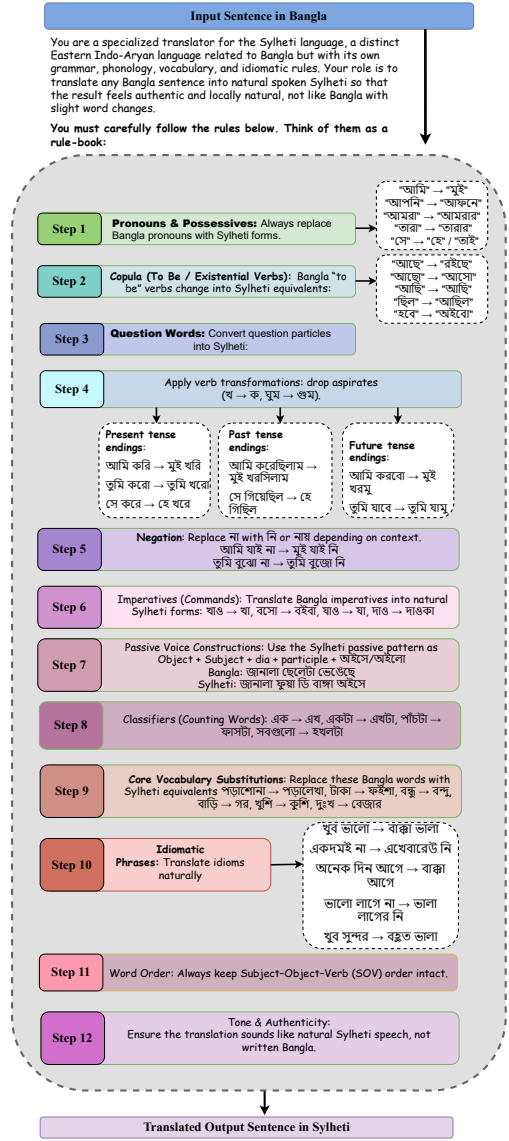

Figure 1: Overview of the Sylheti-CAP prompting framework. The framework consists of three key stages: (1) Linguistic Rulebook Integration with Sylheti-specific grammatical and morphological rules (2) Bilingual Lexicon and Idiom dictionary and (3) Authenticity and Fluency Check.

## 2 The Sylheti-CAP Framework: Prompting for Low-Resource Dialectal Translation

Prompting language models (LMs) for translation, particularly between standard and dialectal variants, assumes that the model has been pre-trained on sufficient parallel data in both languages. For low-resource languages like Sylheti, a dialect of Eastern Indo-Aryan Bangla with distinct phonology, grammar, and vocabulary, this assumption often fails—even in large multilingual LMs. Moreover, translation quality typically declines when faced with out-of-domain data (Zhang and Zong, 2016; Koehn and Knowles, 2017). To address these challenges of data scarcity and domain mismatch, we introduce the Sylheti-CAP (Sylheti Context-Aware Prompting) framework. This method leverages the in-context learning ability of LMs by injecting structured linguistic rules and bilingual lexicons directly into the translation prompt (Figure 1).

Dictionaries and rulebooks are often available even for low-resource languages, making them cost-effective sources of translation knowledge (Arthur et al., 2016; Zhong and Chiang, 2020; Hämäläinen and Alnajjar, 2019). The Sylheti-CAP framework integrates this information into the prompt through a three-part schema to ensure that outputs reflect authentic Sylheti usage rather than slightly modified Bangla.

**Step 1. Linguistic Rulebook.** This section defines the translator persona and the grammatical and phonological rules required for authentic Sylheti output. Key rules include:

- **Pronoun and Possessive Substitution:** আমি (I) → মুই (I), আমার (my) → মোর (my).
- **Copula and Existential Verbs:** আছে (is/are, exists) → রইছে (is/are, exists), আছো (you are) → আসো (you are).
- **Verb Transformations:** Apply phonological simplification (e.g., খ (kh sound) → ক (k sound)) and tense-specific conjugations (আমি করবো (I will do) → মুই খরমু (I will do)).
- **Syntactic and Morphological Directives:** Enforce negation (না (not) → নি/নায় (not)), imperatives (খাও (eat!) → খা (eat!)), and maintain SOV (Subject–Object–Verb) word order.

**Step 2. Core Vocabulary and Idioms Dictionary.** This section provides a lexicon of frequently used words and idiomatic expressions where direct translation is insufficient. It guides the model toward contextually appropriate substitutions and handles non-standard lexical gaps. A dictionary of 3,106 word pairs was created for this purpose (see Appendix A.1).

**Core Vocabulary Examples:**

| Bangla Word | Sylheti |
|---|---|
| পড়াশোনা (study) | পড়ালেখা |
| টাকা (money) | ফইশা |
| বাড়ি (house/home) | গর |
| খুশি (happy) | কুশি |
| বন্ধু (friend) | বন্দু |

**Idiomatic Expressions:**

| Bangla Expression | Sylheti |
|---|---|
| খুব ভালো (very good) | বাক্কা ভালা |
| একদমই না (not at all) | এখেবারেউ নি |
| অনেক দিন আগে (a long time ago) | বাক্কা আগে |
| ভালো লাগে না (do not like / does not feel good) | ভালা লাগের নি |

**Step 3. Sentence-Level Translation and Authenticity Check.** The final segment presents the Bangla source sentence, followed by meta-instructions guiding the model to prioritize fluency and natural spoken style over literal translation. This ensures the generated text reflects authentic Sylheti speech rather than formalized Bangla.

Overall, Sylheti-CAP combines linguistic rules and bilingual dictionaries within a structured prompt, providing an interpretable and adaptable method for high-quality dialect-specific machine translation—especially valuable for under-resourced language pairs where traditional neural MT systems fail to capture dialectal nuances. Appendix A.2 Table 12 shows the prompt we used following Sylheti-CAP framework.

## 3 Experiments

**Dataset.** For evaluation, we use the Vashantor corpus (Faria et al., 2023), which contains 2,500 Sylheti ⇔ Bangla parallel sentences collected from websites, social media platforms, and discussion boards. Each sentence has been professionally translated into Bangla. We use a 375-sentence test set to evaluate each model.

**Dictionaries.** For translation, we employ ground-truth bilingual dictionaries constructed from three Sylheti⇔Bangla parallel datasets: Vashantor (Faria et al., 2023) (2,125 sentences), ONUBAD (Sultana et al., 2025) (980 sentences), and a Sylheti dataset (Prama and Oni, 2025) (5,002 sentence pairs). From these sources, we derived word-level mappings by taking the union of unique tokens, resulting in 2260 distinct words that differ between the Sylheti and Bangla sides. Examples of Sylheti⇔Bangla word mappings are provided in Appendix A.1 (Table 7). Additionally, a large number of words are identical in both languages since Sylheti is a dialect of Bangla (see Table 8).

**Models.** We evaluated five state-of-the-art LLMs from major developers, each with distinct technical specifications. The selection prioritized cutting-edge, diverse architectures to enable a comprehensive competitive assessment.

*LLaMA-4* (AI, 2024) comes in two variants: Llama 4 Scout (17B active parameters, 16 experts) and Llama 4 Maverick (17B active parameters, 128 experts). Llama 4 Maverick is considered the leading multimodal model in its class, outperforming GPT-4o, Gemini 2.0 Flash, and DeepSeek V3 on reasoning and coding benchmarks. In our experiments, we evaluate the Llama 4 Maverick model via the Meta.AI [1] website.

*Gemini 2.5 Flash* (DeepMind, 2025), (released on June 17, 2025) is Google's latest sparse mixture-of-experts Transformer model, optimized for large-context processing with up to 1,048,576 input tokens and 65,535 output tokens. It features advanced reasoning, agentic behaviors, and real-time application support. In this experiment, we evaluated Gemini 2.5 Flash using the Google AI Studio [2] platform.

*GPT-4.1* (OpenAI, 2024) is a multimodal LLM that achieves human-level performance on diverse professional and academic benchmarks. Based on the Transformer architecture, it is pre-trained for next-token prediction and can process up to 32,768 tokens per input. The model is accessible via ChatGPT Plus and the OpenAI API; in this experiment, we accessed and evaluated GPT-4.1 through the OpenAI API.

*Grok 3* (released February 17, 2025) (xAI, 2025) is xAI's latest 1.2-trillion-parameter model, combining transformer-based language modeling with symbolic reasoning modules (Inaba et al., 2003). It uses 128 expert networks with dynamic routing and cross-expert attention gates, achieving 83% parameter activation efficiency while enabling knowledge sharing between experts (Doshi et al., 2023) which is trained on 13.4 trillion tokens. In this experiment, we evaluated Grok 3 through its official web interface [3].

*DeepSeek-V3* (released December 26, 2024) is a Mixture-of-Experts language model with 671 billion total parameters, 37 billion of which are active per token. It employs Multi-head Latent Attention (MLA) and the DeepSeekMoE architecture, extending DeepSeek-V2 for more efficient inference and cost-effective training. Pre-trained on 14.8 trillion tokens and further optimized via supervised finetuning and reinforcement learning. In this experiment, we evaluated DeepSeek-V3 using the official website [4].

**Metrics.** We evaluate LLM performance using BLEU (Bilingual Evaluation Understudy) (Papineni et al., 2002) and ChrF (Character-level F-score) (Popovic, 2015), which together offer a complementary view across tokenization granularities. In addition, we report METEOR (Banerjee and Lavie, 2005), which mitigates some semantic-matching limitations of BLEU by incorporating stemming and synonymy. Taken together, BLEU, ChrF, and METEOR provide a multi-dimensional assessment of translation quality.

**Comparative Methods.** We consider the following prompting strategies:

*Zero-shot.* A direct translation prompt with the model's default settings; temperature is set to 1 in all experiments.

*Few-shot.* In-context learning with exemplars included in the prompt (Hendy et al., 2023). Prior work shows that example selection strategy and count can affect performance (Agrawal et al., 2022; Zhu et al., 2023), with random selection often performing best (Zhu et al., 2023). As the number of examples increases from 1 to 8, BLEU typically improves (Zhu et al., 2023). We use five exemplars in our prompts.

*Chain-of-Thought (CoT).* CoT prompting decomposes translation into structured sub-steps, encouraging the model to reason through lexical, grammatical, and topical aspects before producing the final output (Wei et al., 2022). This approach is inspired by professional human translation workflows (Baker, 1992; Koehn, 2009; Bowker, 2002; Hatim and Munday, 2005).

Appendix A.2 presents the exact prompts used for the four strategies.

---

[1] https://ai.meta.com/
[2] https://deepmind.google/
[3] https://grok.com/
[4] https://deepseekv3.org/

# 4 Results and Discussion

## 4.1 RQ1: Benchmarking LLMs for Sylheti ⇔ Bangla

Figure 2 shows the BLEU score in both translation directions (Sylheti ⇔ Bangla). Across both directions, Grok 3 and LLaMA 4 are the strongest models, with LLaMA 4 leading Bangla→Sylheti (BLEU-1 = 0.3565; Grok 3 = 0.3525) and Grok 3 leading Sylheti → Bangla (0.4855; LLaMA 4 = 0.4656), while GPT-4.1 and Deepseek V3.2 trail on Bangla → Sylheti (both 0.2106). A pronounced directional asymmetry emerges: every model performs substantially better when translating into Bangla than into Sylheti—for the top systems, Grok 3 is 1.38 times higher (0.4855 vs. 0.3525) and LLaMA 4 is 1.31 times higher (0.4656 vs. 0.3565) on Sylheti → Bangla than Bangla → Sylheti, indicating that current LLMs are more proficient at producing the high-resource standard language than generating the dialect. This gap likely stems from pre-training data imbalance and limited exposure to Sylheti's lexicon, morphology, and orthography; as a result, models often normalize dialectal items into standard Bangla or omit Sylheti-specific function words. Qualitative examples in Table 1 show that zero-shot LLMs normalize Sylheti into standard Bangla, erasing dialectal lexicon, morphology, and particles. Core Sylheti words, e.g., ফুড়িটা (the girl), এখইন (now), যাইবা (will go), ফুয়াটায় (the boy), ফারলো (could/was able to), বাফর (father), মাই (mother), কেনিয়া (having bought), আনছইন (has brought), and আছইন নি? (is he not well?) are replaced by Bangla-leaning forms like মেয়াডা (the girl), এখন (now), যা (go), পোলাডা (the boy), পারল (could/was able to), আব্বার (father's), আম্মা (mother), কিনা (having bought), আনছে (has brought), and কেমন আছেগো? (how are you?). These errors reflect lexical substitution, morphological normalization (future, negation, honorifics), and orthographic drift, indicating limited Sylheti exposure and a decoding prior biased toward standard Bangla.

## 4.2 RQ2: Enhance LLM's Translation Performance by Sylheti-CAP

The evaluation of prompting strategies for both Bangla ⇔ Sylheti translation tasks across five LLMs shows a clear and consistent advantage for the proposed Sylheti-Context-Aware Prompting (Sylheti-CAP) method. Table 2 and 3 shows Sylheti-CAP achieves the highest scores across

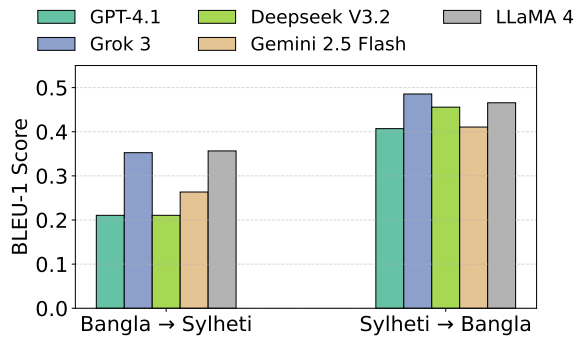

Figure 2: BLEU-1 scores on the test dataset for five LLMs (GPT-4.1, GPT-4.1-mini, LLaMA 4, Grok 3, and Deepseek V3.2) evaluated in both Bangla ⇔ Sylheti translation directions. BLEU scores are averaged over all test samples in each translation direction for this experiment.

all models and evaluation metrics (BLEU-1, METEOR, and ChrF) in both translation directions.

As shown in Table 2, Sylheti-CAP consistently outperforms Zero-Shot, Few-Shot, and Chain-of-Thought (CoT) prompting. For example, Grok achieved the highest BLEU-1 (0.47) and ChrF (46.01), improving significantly over its Zero-Shot baseline (0.35 BLEU-1, 42.19 ChrF). LLaMA and GPT attained the top METEOR score of 0.34, while Deepseek's ChrF rose from 35.81 to 39.07 and Gemini improved from 0.26 to 0.29 BLEU-1. These gains highlight Sylheti-CAP's ability to inject dialect-specific context and structure into LLMs, enhancing lexical and semantic accuracy even in low-resource conditions.

Similarly, Table 3 demonstrates that Sylheti-CAP generalizes effectively in the reverse direction. Across all LLMs, it again delivers the best results for every metric. Deepseek reached the highest BLEU-1 (0.52), METEOR (0.45), and ChrF (53.72), outperforming its next-best CoT configuration. Grok followed closely with 0.52 BLEU-1 and 53.13 ChrF, while LLaMA, GPT, and Gemini also recorded steady improvements. For instance, LLaMA's ChrF increased from 49.95 to 52.19, and GPT's METEOR rose from 0.34 to 0.43, demonstrating Sylheti-CAP's consistent enhancement of lexical precision and semantic alignment. Across both translation directions, Sylheti-CAP proves highly effective in addressing data scarcity and capturing dialectal nuance. In contrast, Zero-Shot and CoT rely on general multilingual reasoning and lack sensitivity to dialectal variation. As shown in Table 4, the baseline prompts frequently fall back to standard Bangla forms or non-target dialect words (e.g., বাপে (father), বাফা (father), আফা (older sister), পর (after), পরত (after/afterwards))), whereas Sylheti-CAP reliably selects the intended Sylheti lexicon such as আব্বা (father), ফরে (after), and বইনের (sister's). Only Sylheti-CAP yields a near-target variant, while other prompts produce lexically and morphologically off-target variants like বাপে কিরাম (how is your father) or কেমন আছের (how are you), and পর/পরত (after), but Sylheti-CAP correctly preserves core Sylheti dialect-specific word-to-word mappings for achieving lexically faithful Bangla→Sylheti translations.

Table 1: Examples of Standard Bangla, Original Sylheti, and LLM-generated Sylheti translations produced by LLaMA 4 under the default prompt setting. Words highlighted in green represent accurate Sylheti translations corresponding to the Bangla source, while red highlights indicate errors or hallucinated terms generated by the model during Bangla→Sylheti translation.

| Bangla Sentence | Original Sylheti | Translated Sylheti |
| --- | --- | --- |
| তোমার আব্বু কেমন আছে? (How is your father?) | তোমার আব্বা বালা আছইন নি? | তোর আব্বা কেমন আছে-গো? |
| মেয়েটি এখন সিলেট যাবে (The girl will now go to Sylhet) | ফুড়িটা এখইন সিলট যাইবা | মেয়াডা এখন সিলট যা |
| ছেলেটি আমাকে চিনতে পারল না (The boy could not recognize me) | ফুয়াটায় আমারে চিনতে ফারলো না | পোলাডা আমারে চিনবার পারল নাইব |
| আমার আব্বুর খুব গরম লাগছে (My father is feeling very hot) | আমার বাফর খুব গরম লাগছে | আমার আব্বার অনেক গরম লাগতাছে |
| আম্মু আমার জন্য নীল শাড়ি কিনে এনেছে(My mom bought a blue sari for me) | মাই আমার লাগি নীল শাড়ি কেনিয়া আনছইন | আম্মা আমার লাগি নীল শাড়ি কিনা আনছে |

### 4.3 Human Evaluation.

We conducted a human preference study on 200 samples for the Bangla ⇔ Sylheti translation task with 3 native speakers in Sylheti. Annotators rated translations from four prompting strategies Zero-Shot, Few-Shot, CoT, and Sylheti-CAP as Good, Fair, or Poor. Figure 3 shows that Sylheti-CAP consistently achieved the highest proportion of Good translations in both directions. For instance, Deepseek V3.2 and Grok 3 reached over 50% Good ratings in both Bangla → Sylheti and Sylheti → Bangla, while Poor outputs stayed below 20%. Overall, Sylheti-CAP substantially reduced low-quality outputs and increased human preference, confirming its effectiveness for dialect-aware translation.

### 4.4 LLM-as-a-judge.

We also conducted an LLM-as-a-judge study on the same set of 200 samples used in the human evaluation for the Bangla ⇔ Sylheti translation task. Using GPT-5.0, we directly scored adequacy, fluency, and overall translation quality on a 0–100 scale by comparing the reference Sylheti sentence with LLM-generated Sylheti translations under different prompting strategies. Appendix A.2 and Table 13 present the prompt used in the LLM-as-a-judge setup. Table 5 shows that Sylheti-CAP consistently achieves the highest adequacy, fluency, and overall scores, outperforming all other prompting strategies by a margin of 3–10 points.

### 4.5 MQM Evaluation.

To further analyze translation quality improvements across prompting strategies, we conducted Multidimensional Quality Metric (MQM) evaluations (Lommel, 2013) using the same 200 samples from the Bangla ⇔ Sylheti test sets. Following the expert-based annotation protocols in (Freitag et al., 2021; He et al., 2023), annotators identified translation errors, categorized them (e.g., omission, untranslated text, awkward phrasing, and mistranslation), and rated their severity. Each category contributed a weighted penalty, producing an overall MQM score per system.

As summarized in Table 6, Sylheti-CAP achieved the lowest (best) MQM scores in both directions (1.62 for Ben→Syl and 1.93 for Syl→Ben), outperforming Zero-Shot, Few-Shot, and CoT prompting. The category-level breakdown in Figure 4 shows that these improvements are primarily driven by reductions in mistranslations, awkward phrasing, and omission errors, where Sylheti-CAP consistently yields lower penalties (e.g., 580 vs. 670 for mistranslation and 200 vs. 220 for omission compared to Zero-Shot). These findings indicate that incorporating dialectal context and linguistic grounding not only reduces literal translation errors but also enhances overall fluency and semantic adequacy.

Table 2: Translation performance (BLEU, ChrF, METEOR) of GPT-4.1, GPT-4.1-mini, LLaMA 4, Grok 3, and Deepseek V3.2 for ``Bangla→Sylheti'' translation. Scores are the average of each test set for each language, measured using BLEU, ChrF, and METEOR metrics. Orange shading indicates that Sylheti-CAP outperformed other prompt strategies.

| Model | Zero-Shot | | | Few-Shot | | | COT | | | Sylheti-CAP | | |
|---|---|---|---|---|---|---|---|---|---|---|---|---|
| | Bl | M | C | Bl | M | C | Bl | M | C | Bl | M | C |
| **Deepseek** | 0.21 | 0.24 | 35.81 | 0.10 | 0.07 | 14.47 | 0.27 | 0.19 | 35.38 | 0.32 | 0.24 | 39.07 |
| **Grok** | 0.35 | 0.28 | 42.19 | 0.39 | 0.27 | 41.57 | 0.33 | 0.26 | 39.81 | 0.47 | 0.30 | 46.01 |
| **LLaMA** | 0.36 | 0.26 | 37.09 | 0.35 | 0.32 | 42.22 | 0.34 | 0.25 | 38.23 | 0.42 | 0.34 | 45.08 |
| **GPT** | 0.36 | 0.32 | 42.68 | 0.32 | 0.30 | 43.34 | 0.34 | 0.29 | 40.60 | 0.42 | 0.34 | 43.91 |
| **Gemini** | 0.26 | 0.19 | 34.71 | 0.23 | 0.15 | 30.51 | 0.19 | 0.14 | 31.61 | 0.29 | 0.24 | 35.86 |

Table 3: Translation performance (BLEU, ChrF, METEOR) of GPT-4.1, GPT-4.1-mini, LLaMA 4, Grok 3, and Deepseek V3.2 for ``Sylheti→Bangla'' translation. Scores are the average of each test set for each language, measured using BLEU, ChrF, and METEOR metrics. Blue shading indicates that Sylheti-CAP outperformed other prompt strategies..

| Model | Zero-Shot | | | Few-Shot | | | COT | | | Sylheti-CAP | | |
|---|---|---|---|---|---|---|---|---|---|---|---|---|
| | Bl | M | C | Bl | M | C | Bl | M | C | Bl | M | C |
| **Deepseek** | 0.46 | 0.39 | 51.61 | 0.44 | 0.37 | 49.98 | 0.50 | 0.42 | 51.23 | 0.52 | 0.45 | 53.72 |
| **Grok 3** | 0.49 | 0.41 | 49.92 | 0.49 | 0.41 | 49.54 | 0.47 | 0.39 | 48.11 | 0.52 | 0.44 | 53.13 |
| **LLaMA** | 0.47 | 0.41 | 49.95 | 0.45 | 0.37 | 47.73 | 0.45 | 0.39 | 51.01 | 0.49 | 0.41 | 52.19 |
| **GPT** | 0.41 | 0.34 | 46.45 | 0.50 | 0.40 | 48.35 | 0.41 | 0.33 | 44.09 | 0.47 | 0.43 | 51.49 |
| **Gemini** | 0.41 | 0.34 | 46.72 | 0.41 | 0.34 | 46.83 | 0.40 | 0.33 | 45.95 | 0.46 | 0.39 | 48.69 |

### 4.6 LLMs' Hallucinations.

In natural language generation (NLG), hallucination refers to the production of content that is nonsensical or unfaithful to the source text (Filippova, 2020; Zhang et al., 2019), and remains a persistent challenge for LLMs (Zhang et al., 2023). To examine this issue within the context of Bangla ⇔ Sylheti translation, we conducted a human evaluation of hallucination errors across four prompting strategies. Using 200 sampled sentences from each translation direction, annotators inspected the generated outputs from five LLMs and labeled whether each contained hallucinated or semantically inconsistent content, following the definition in (Guerreiro et al., 2023).

As illustrated in Figure 5, Sylheti-CAP consistently achieves the lowest hallucination rates across all models (e.g., 12.6–13.8%), outperforming CoT, Few-Shot, and Zero-Shot prompting, which exhibit higher rates (typically 15–17%). We attribute this reduction to the contextual grounding of Sylheti-CAP, which integrates dialect-specific translation cues and semantic constraints directly into the prompt. This additional linguistic guidance helps steer the model's token generation away from spurious continuations, improving overall faithfulness and reducing nonsensical or unaligned outputs.

## 5 Related Works

**LLMs in Machine Translation.** Recent advances in LLMs such as GPT-4 (OpenAI et al., 2023) and LLaMA (Touvron et al., 2023) have significantly advanced Neural Machine Translation (NMT) (Jiao et al., 2023; Hendy et al., 2023). Two main paradigms dominate: in-context learning (ICL) and fine-tuning. ICL enables LLMs to perform translation tasks from a few exemplars without parameter updates (Brown et al., 2020), often matching supervised models (García et al., 2023). The quality of demonstrations strongly influences performance (Agrawal et al., 2022). In contrast, fine-tuned models such as XGLM-7B (Li et al., 2023) and instruction-tuned variants (Chen et al., 2021) improve translation faithfulness and low-resource adaptability.

Table 4: Examples of Standard Bangla, Original Sylheti, and LLM-generated Sylheti translations produced by LLaMA 4 under the Zero-Shot Few-Shot COT Sylheti-CAP prompt setting. Words highlighted in green represent accurate Sylheti translations corresponding to the Bangla source, while red highlights indicate errors or hallucinated terms generated by the model during Bangla→Sylheti translation.

| Bangla | Sylheti | Zero-Shot | Few-Shot | COT | Sylheti-CAP |
|---|---|---|---|---|---|
| তোমার আব্বু কেমন আছে? | তোমার আব্বা বালা আছইন নি? | তোমার বাপে কিরাম আছে? | তোমার আব্বা কেমন আছের? | তুমার বাফা ক্যামন আছইন? | তুমার আব্বা বালা আছইন? |
| আমার দুইদিন পরে বিয়ে হবে | আমার দুইদিন ফরে বিয়া ই-বো | আমার দুই দিন পরত বিয়া অইব | আমার দুই দিন পর বিয়া অইবো | আমার দুই দিন পর বিয়া অইব | আমার দুই দিন ফরে বিয়া অইবো |
| আমার বড় বোনের আজকে মন ভালো নেই | আমার বড় বইনর আইজ মন ভালা নায় | আমার বড় আফা অহন মন বালা নাই | আমার বড় আফার আইজকু মন ভালা নায় | আমার বড় আফার আজকা মন বালা নাই | আমার বড় বইনের আইজকু মন ভালা নায় |

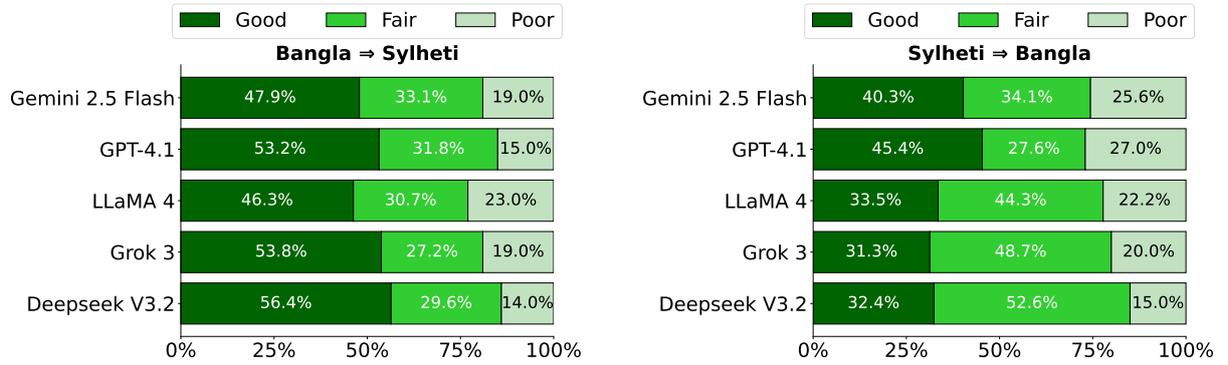

Figure 3: Human preference study comparing Sylheti-CAP with Zero-Shot, Few-Shot and COT for LLMs (GPT-4.1, GPT-4.1-mini, LLaMA 4, Grok 3, and Deepseek V3.2).

Table 5: GPT-5.1-as-a-judge average scores (0–100) for adequacy, fluency, and overall translation quality, comparing reference Sylheti sentences with LLM-generated Sylheti translations across different prompting strategies.

| Prompt | Adequacy | Fluency | Overall |
|---|---|---|---|
| Zero-shot | 72.7 | 77.5 | 75.6 |
| Few-shot | 78.5 | 79.5 | 82.4 |
| CoT | 76.3 | 78.2 | 78.8 |
| **Sylheti-CAP** | **84.2** | **86.5** | **85.3** |

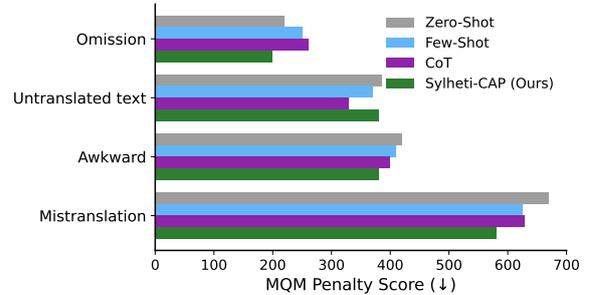

Figure 4: MQM penalty scores across different error categories for 200 test sentences from each of the Bangla⇔Sylheti test sets. Lower scores is less severe translation errors.

Evaluation of LLM-based translation generally follows two directions: (1) Prompt-level design, focusing on prompt templates, demonstration selection, and reasoning structure (Vilar et al., 2022; Zhang et al., 2023; Jiao et al., 2023); and (2) Comprehensive benchmarking, testing multilingual (Hendy et al., 2023; Zhu et al., 2023), document-level (Karpinska and Iyyer, 2023), low-resource (García et al., 2023), and hallucination-resistant (Guerreiro et al., 2023) settings, often incorporating human feedback (Jiao et al., 2023). While early efforts to use cross-sentence context showed limited gains (Lopes et al., 2020; Fernandes et al., 2021), recent LLMs can dynamically leverage document-level and contextual cues (Karpinska and Iyyer, 2023; Wang et al., 2023). Newer methods integrate retrieval-based prompting (Agrawal et al., 2022), bilingual lexicons (Ghazvininejad et al., 2023), context-aware prompting (Pilault et al., 2023), and document-level fine-tuning (Wu et al., 2024). However, LLMs' potential to fully exploit bilingual, multi-turn contextual signals and context-aware evaluation remains underexplored—particularly for low-resource and dialectal translation, where context injection can close significant linguistic gaps.

Table 6: Averaged MQM scores (↓) for different prompting strategies on Bangla–Sylheti (Ben→Syl) and Sylheti–Bangla (Syl→Ben) translation tasks. Lower values indicate fewer translation errors and better quality.

| Prompt | Ben→Syl | Syl→Ben |
| --- | --- | --- |
| Zero-Shot | 2.54 | 3.02 |
| Few-Shot | 2.41 | 2.87 |
| CoT | 2.18 | 2.56 |
| Sylheti-CAP | **1.62** | **1.93** |

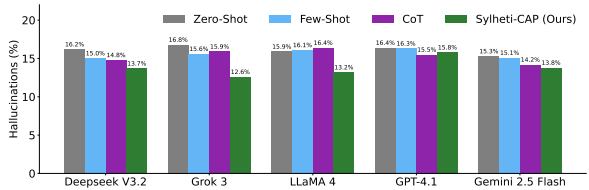

Figure 5: Ratio of hallucinations in generated translations for 200 test sentences from each of the English⇔Bengali test sets. Human annotators labeled each output as either containing or not containing a hallucination error.

**Bangla Machine Translation.** Early MT efforts for Bangla concentrated on the high-resource Bangla–English pair. For Bangla → English, studies have employed Sequence-to-Sequence (Seq-to-Seq) models utilizing attention-based Recurrent Neural Networks (RNNs) (Islam et al., 2023). Conversely, English → Bangla translation has been successfully achieved using encoder–decoder Gated Recurrent Unit (GRU) architectures, which were shown to outperform LSTM-based models (Mahmud et al., 2021). Beyond specific models, comprehensive analyses have benchmarked multiple NMT architectures for the general Bangla–English task (Hasan et al., 2019). More recent work has leveraged transformer-based models with large-scale multi-dialect parallel corpora to address generalized dialectal Bangla translation (Faria et al., 2023). Addressing dialectal variation, efforts on the Chittagonian dialect have applied rule-based morphological transformations and bidirectional mappings for conversion (Milon et al., 2020; Hossain et al., 2022). For Sylheti, foundational work has provided essential grammatical insights (Goswami, 2021). In NMT, a Sylheti → Bangla system was previously introduced using a BiLSTM (Prama and Anwar, 2025b) and transformer based architecture (Oni and Prama, 2025). Despite these contributions, the Bangla–Sylheti pair remains significantly underexplored due to scarce standardized corpora, substantial orthographic variation, and limited linguistic resources. To the best of our knowledge, this study is the first to employ and systematically evaluate Large Language Models (LLMs) for the challenging Bangla ⇔ Sylheti dialect translation task.

## 6 Conclusion

This study presents the first systematic evaluation of Large Language Models (LLMs) for Bangla–Sylheti Machine Translation. We propose Sylheti-CAP (Context-Aware Prompting), a framework that integrates linguistic rules, bilingual dictionaries, and contextual fluency constraints directly into prompts to generate accurate and natural Sylheti translations. Experiments across five advanced LLMs (GPT-4.1, GPT-4.1-mini, LLaMA 4, Grok 3, and Deepseek V3.2) show that Grok 3 and LLaMA 4 achieve the highest BLEU and METEOR scores in both translation directions. Sylheti-CAP consistently outperforms zero-shot, few-shot, and chain-of-thought baselines, reducing hallucinations, mistranslations, and awkward phrasing. Overall, Sylheti-CAP demonstrates a scalable, linguistically grounded approach for low-resource and dialectal translation, paving the way for improved translation quality across other Bangla dialects and underrepresented languages.

## 7 Limitations

While Sylheti-CAP demonstrates significant improvements in Bangla–Sylheti translation, several limitations remain. The framework relies solely on prompting without model fine-tuning. Incorporating fine-tuned word embeddings could provide a more stable and permanent improvement in translation performance. Current bilingual dictionary consists of only 2260 word pairs expanding it to include a wider range of dialect-specific and context-rich words would likely enhance translation quality and coverage. Prominent LLMs used in this study—such as GPT-4.1, LLaMA 4, and Deepseek V3.2 are primarily trained on data from high-resource languages. Since their pretraining corpora likely contain limited or no Sylheti text, this lack of exposure may constrain their dialectal understanding. Moreover, the absence of publicly available training data for proprietary models limits its reproducibility and transparency. Finally, our human evaluation involved a small number of native Sylheti speakers from different regions. Although care was taken to ensure linguistic proficiency and regional diversity, subjective variation remains, and the results may not fully generalize. Conducting broader evaluations with more participants and developing standardized Sylheti evaluation datasets would strengthen benchmarking and comparability in future work.

Table 7: Examples from the Bangla⇔Sylheti word-to-word dictionary.

| Bangla | Sylheti |
|---|---|
| মহিলার | বেটির |
| হবে | অইবো |
| উপরে | উফরে |
| একটাই | এখটাউ |
| একেক | এখনত |
| এলাকার | জাগার |
| রকম | লাখান |
| নতুনরা | নয়া |
| অনেক | বহুতত |
| শিখতে | হিকতা |
| করি | খরি |
| হোকনা | অউক |
| কিছুর | কুন্তার |
| শুরুটা | শুরু |
| এভাবেই | অলাউ |
| আমার | মোর |
| সাথে | লগে |
| কথা | মাতাবায় |
| অনেক | বাক্কা |
| হবে | লাগবো |
| সবার | হখলর |
| কত | খত |

# A  Appendix

## A.1  Bangla⇔Sylheti Dictionary

To build a comprehensive bilingual lexicon, we merged three parallel corpora: Vashantor (Faria et al., 2023) (2,125 sentences), ONUBAD (Sultana et al., 2025) (980 sentences), and the Sylheti Dataset (Prama and Oni, 2025) (5,002 sentence pairs). Since Sylheti is a dialect of Bangla, a large portion of the vocabulary overlaps between the two. However, there are also numerous dialect-specific variations in phonology, morphology, and semantics. From these datasets, we compiled a word-to-word dictionary containing 2260 aligned sentence pairs, focusing on words and expressions unique to Sylheti. Here is the dictionary of Bangla⇔Sylheti Dictionary: https://github.com/word mapping 2260.csv. Table 7 shows the Bangla⇔Sylheti word-to-word dictionary.

## A.2  Prompt Strategies

Table 8: Examples of identical words in Bangla and Sylheti. While Sylheti is a dialect of Bangla, many words remain unchanged due to shared linguistic roots, phonetic overlap, and common Indo-Aryan origin. These lexical similarities contribute to overall translation fluency between the two languages.

| Bangla | Sylheti |
|---|---|
| তুমি | তুমি |
| রাজকুমারির | রাজকুমারির |
| মায়া | মায়া |
| জীবন | জীবন |
| রঙিন | রঙিন |
| ছবি | ছবি |
| আর | আর |
| আশেপাশে | আশেপাশে |

Table 9: Zero-Shot Prompt: Direct instruction for Bangla→Sylheti translation without examples or prior context.

**Prompt:**

You are a professional translator proficient in both Bangla and Sylheti. Your task is to translate the following Bangla sentence into natural and fluent Sylheti. Provide only the translated Sylheti sentence without any additional explanation.
**Bangla:** "<input_sentence>"
**Sylheti:**

Table 10: Few-Shot Prompt: Translation prompt with six Bangla–Sylheti example pairs to guide model behavior.

**Prompt:**

You are given Bangla sentences and asked to translate them into Sylheti. Here are a few examples:
**Bangla–Sylheti Examples:**
১. কেমন আছো ? → ভালা আছনি?
২. আজকে আমার মন ভালো নেই → আইজকু আমার মন ভালা নায়
৩. তুমি কি করো ? → তুমি কিতা খরো?
৪. এই গরমে আমার কিছু ভালো লাগে না → অউ গরমো আমার কুনতা ভালা লাগের না
৫. ছেলেটি সাদা রঙয়ের একটি শার্ট পরে এসেছিল → ফুয়াটায় এখটা সাদা রংগর শার্ট পিন্দিয়া আইছিল

**Instruction:** Translate the following Bangla sentence into Sylheti:
**Bangla:** "<input_sentence>"
**Sylheti:**

Table 11: Chain-of-Thought (CoT) Prompt: A structured, reasoning-based prompt for multi-step contextual translation.

**Prompt:**

You are a translation assistant that follows a three-step process: **Knowledge Mining → Knowledge Integration → Knowledge Selection.** Your goal is to translate the given Bangla text into Sylheti as accurately and fluently as possible.

**Step 1: Knowledge Mining** 1. Extract the keywords from the input **Bangla** sentence and translate them into Sylheti. *Output:* Keyword Pairs: <src_word1>:<tgt_word1>, ...

2. Identify a few words describing the main topics of the sentence. *Output:* Topics: <topic1>, <topic2>, ...

3. Write a **Bangla** sentence related to but different from the input, and provide its **Sylheti** translation. *Output:* <src_demo> | <tgt_demo>

**Step 2: Knowledge Integration** Combine the mined knowledge to generate a candidate translation.

*Prompt:* Keyword Pairs: ...

Topics: ...

Related Example: <src_demo> | <tgt_demo>

Instruction: Given the above, translate the following **Bangla** sentence into **Sylheti**.

**Bangla:** "`<input_sentence>`"

**Sylheti:** <Candidate Translation>

**Step 3: Knowledge Selection** Compare all candidate outputs (Keyword, Topic, Demo, Base) and select the most fluent and accurate final translation.

*Output:* Best Translation: <final_output>

Table 12: Sylheti-CAP Prompt: Context-Aware Prompt integrating explicit linguistic rules and word mappings for authentic Bangla→Sylheti translation.

**Prompt:**

You are a translator specializing in **Sylheti**, a distinct Indo-Aryan language closely related to Bangla but with its own grammar, vocabulary, and phonology. Your task is to translate Bangla sentences into natural, fluent Sylheti speech while preserving meaning, grammar, and idiomatic usage. Follow all the rules and mapping guidelines below when producing the translation.

☐☐ **Grammar and Pronouns:**

- Replace Bangla pronouns with Sylheti equivalents: আমি → মুই, তুমি → তুমি/তুই, আপনি → আফনে, আমরা → আমরার, তারা → তারার, সে → হে/তাই.
- For possessives: আমার → মোর, তোমার → তুমার, আমাদের → আমরার, আপনাদের → আফনারার.

☐☐ **Questions:** Use Sylheti interrogatives. কী → কিতা, কোথায় → কুনান/কুনানো, কেমন → কিলা, কেন → কিয়েন, কত → কিত্তা.

☐☐ **Verbal Rules:**

- Drop aspiration: খ → ক, ঘুম → গুম.
- Present tense endings: আমি করি → মুই খরি, তুমি করো → তুমি খরো, সে করে → হে খরে.
- Past tense: করেছিলাম → খরসিলাম.
- Future tense: করবো → খরমু.
- Negation: না → নি / নায়. Example: আমি যাই না → মুই যাই নি.
- Copula: আছে / আছি / আছো → রইছে / আছি / আসো.

☐☐ **Vocabulary:** পড়াশোনা → পড়ালেখা, টাকা → ফইশা, বন্ধু → বন্দু, বাড়ি → গর, খুশি → কুশি, দুঃখ → বেজার.

☐☐ **Imperatives:** খাও → খা / খাইওকা (polite), বসো → বইবা, যাও → যা.

☐☐ **Passive Voice:** জানালা ছেলেটা ভেঙেছে → জানালা ফুয়া ডি বাঙ্গা অইসে. Pattern: *Object + Subject + dia + participle + oisil/oise/or*.

☐☐ **Classifiers:** একটা → এখটা, পাঁচটা → ফাসটা.

☐☐ **Syntactic and Morphological Directives:** Always preserve the SOV (Subject–Object–Verb) order. Modify pronouns, verbs, negations, and key vocabulary to reflect Sylheti tone and grammar. Output must sound like spoken Sylheti, not formal Bangla.

☐☐ **Reference Word Mapping Dictionary (Excerpt):** Use the following word-level mappings when applicable: মহিলার → বেটির, হবে → অইবো, এলাকার → জাগার, শিখতে → হিকতা, করি → খরি, ভালো → ভালা, সাথে → লগে, কথা → মাতবায়, ছবি → ছবি, যাবে → যাইবো, কিছু → কুনতা, আমার → মোর, আপনি → আফনে.

**Final Instruction:** Translate the following Bangla text into fluent Sylheti, adhering to all rules and mappings above. Ensure the translation reflects natural spoken Sylheti and not literal Bangla.

**Bangla:** `"<input_sentence>"`
**Sylheti:**

Table 13: Sylheti-CAP Prompt: Context-Aware Prompt integrating explicit linguistic rules and word mappings for authentic Bangla→Sylheti translation.

**Prompt:**

**LLM-as-a-judge prompt**

You are an expert bilingual evaluator.

Your task is to evaluate a MACHINE TRANSLATION from Standard Bangla to Sylheti.

SOURCE (Standard Bangla): <*SOURCE SENTENCE*>

REFERENCE TRANSLATION (Sylheti): <*REFERENCE TRANSLATION*>

CANDIDATE TRANSLATION (Translated Sylheti): <*CANDIDATE TRANSLATION using different prompt strategy*>

Please rate the candidate translation on a scale from 0 to 100 for:

1. ADEQUACY: how well it preserves the meaning of the source.

2. FLUENCY: how natural and grammatically correct the text is in Sylheti.

3. OVERALL: your overall judgment of translation quality.

Return your answer in JSON format ONLY, as:

```
{"adequacy": X, "fluency": Y, "overall": Z}
```